\documentclass[conference]{IEEEtran}

\IEEEoverridecommandlockouts
\usepackage{cite}
\usepackage{amsmath,amssymb,amsfonts}
\usepackage{algorithmic}
\usepackage{graphicx}
\usepackage{textcomp}
\usepackage{xcolor}
\def\BibTeX{{\rm B\kern-.05em{\sc i\kern-.025em b}\kern-.08em
    T\kern-.1667em\lower.7ex\hbox{E}\kern-.125emX}}
    
\usepackage{multirow}
\usepackage{float}
\newcommand{\etal}{\textit{et al}. }
\newcommand{\ie}{\textit{i}.\textit{e}. }
\newcommand{\eg}{\textit{e}.\textit{g}. } 
\usepackage{color}
\IEEEoverridecommandlockouts
\usepackage{marvosym}
\usepackage{fancyhdr}

\begin{document}

\title{More than Encoder: Introducing Transformer Decoder to Upsample\\
}


\DeclareRobustCommand*{\IEEEauthorrefmark}[1]{%
  \raisebox{0pt}[0pt][0pt]{\textsuperscript{\footnotesize #1}}%
}

\author{
    \IEEEauthorblockN{
        Yijiang Li\IEEEauthorrefmark{1,2}, 
        Wentian Cai\IEEEauthorrefmark{1,2}, 
        Ying Gao\IEEEauthorrefmark{1,2,\Letter} \thanks{\textsuperscript{\Letter} Corresponding author: Ying Gao, gaoying@scut.edu.cn}, 
        Chengming Li\IEEEauthorrefmark{3} and
        Xiping Hu\IEEEauthorrefmark{4}
    }
    \IEEEauthorblockA{
        \IEEEauthorrefmark{1}\textit{School of Computer Science and Engineering}, \textit{South China University of Technology}, Guangzhou, China\\
        \IEEEauthorrefmark{2}\textit{Guangdong Provincial Key Laboratory of Artificial Intelligence in Medical Image Analysis and Application}, \\\textit{Guangdong Provincial People's Hospital}, \textit{Guangdong Academy of Medical Sciences}, Guangzhou, China\\
        \IEEEauthorrefmark{3}\textit{School of Intelligent Systems Engineering}, \textit{Sun Yat-sen University}, Shenzhen, China\\
        \IEEEauthorrefmark{4}\textit{School of Medical Technology}, \textit{Beijing Institute of Technology}, Beijing, China\\
        \{csliyijiang3000, cscaiwentian\}@mail.scut.edu.cn, 
        gaoying@scut.edu.cn, 
        lichengming@mail.sysu.edu.cn,
        huxp@bit.edu.cn
    }
}

\maketitle

\thispagestyle{fancy}
\lfoot{Accepted by IEEE BIBM 2022}
\renewcommand{\headrulewidth}{0pt} 
\cfoot{}

\begin{abstract}
Medical image segmentation methods downsample images for feature extraction and then upsample them to restore resolution for pixel-level predictions. In such schema, upsample technique is vital in restoring information for better performance. However, existing upsample techniques leverage little information from downsampling paths. The local and detailed feature from the shallower layer such as boundary and tissue texture is particularly more important in medical segmentation compared with natural image segmentation. To this end, we propose a novel upsample approach for medical image segmentation, Window Attention Upsample (WAU), which upsamples features conditioned on local and detailed features from downsampling path in local windows by introducing attention decoders of Transformer. WAU could serve as a general upsample method and be incorporated into any segmentation model that possesses lateral connections. We first propose the Attention Upsample which consists of Attention Decoder (AD) and bilinear upsample. AD leverages pixel-level attention to model long-range dependency and global information for a better upsample. Bilinear upsample is introduced as the residual connection to complement the upsampled features. Moreover, considering the extensive memory and computation cost of pixel-level attention, we further design a window attention scheme to restrict attention computation in local windows instead of the global range. We evaluate our method (WAU) on classic U-Net structure with lateral connections and achieve state-of-the-art performance on Synapse multi-organ segmentation, Medical Segmentation Decathlon (MSD) Brain, and Automatic Cardiac Diagnosis Challenge (ACDC) datasets. We also validate the effectiveness of our method on multiple classic architectures and achieve consistent improvement.
\end{abstract}

\begin{IEEEkeywords}
Transformer, upsampling, semantic segmentation, medical image analysis.
\end{IEEEkeywords}

\begin{figure}[htbp]
\centering
\includegraphics[width=\linewidth]{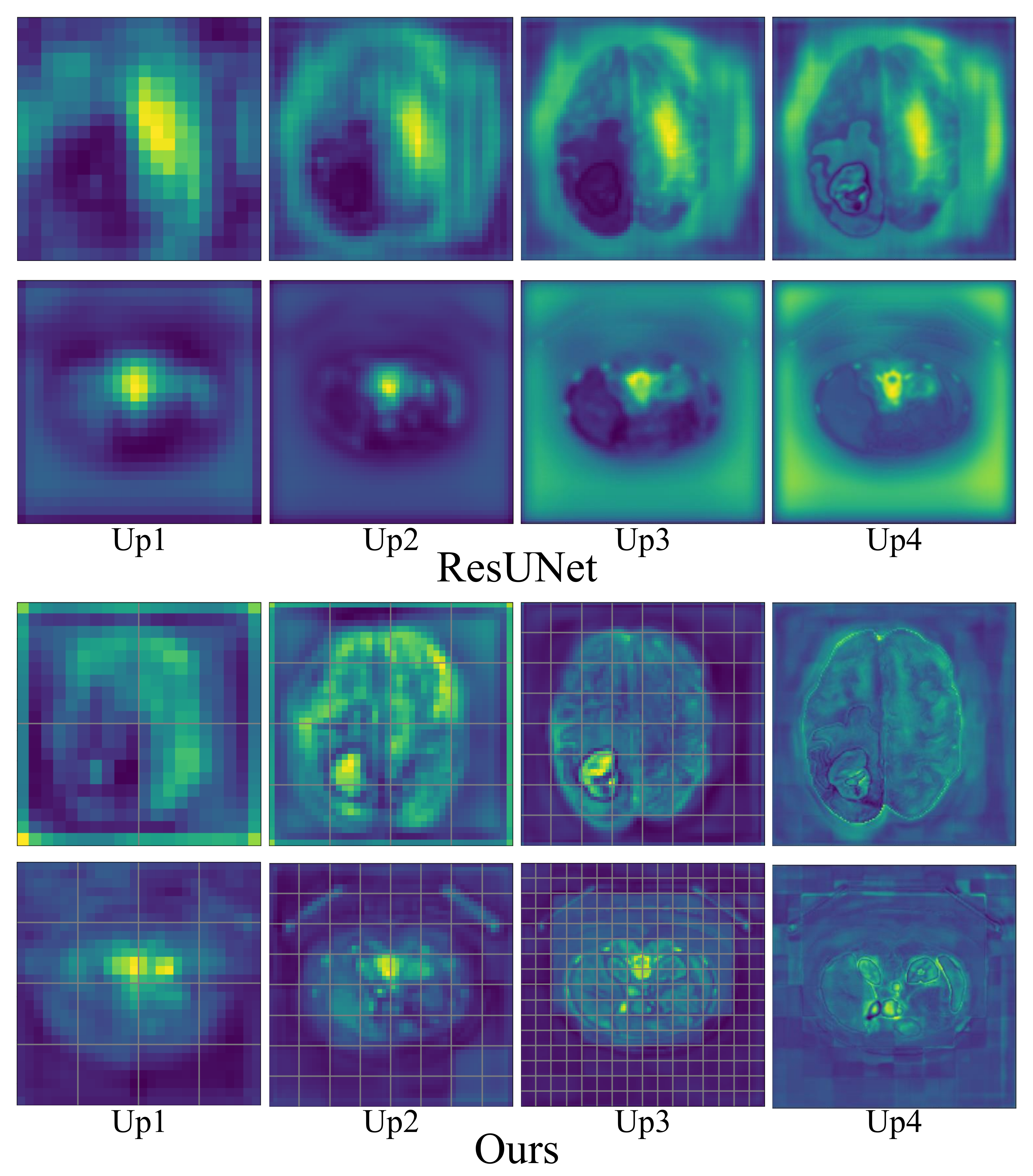}
\caption{Visualization of decoded feature maps during upsampling on MSD Brain and Synapse datasets.}
\label{fig:decoder_feature}
\end{figure}

\section{Introduction}
Deep learning revolutionizes many fields of machine intelligence including multimedia processing \cite{rao2021augmented,9286840,9466418}, scene understanding \cite{wang2022consistent,XIE2022}, and Computer Aided Diagnosis (CAD) \cite{9851919,9855388} area. In CAD area, particularly, medical image segmentation plays a crucial role in clinical diagnosis and treatment processes. Long \etal \cite{long2015fully} proposes the famous FCN architecture which downsamples high resolution images to extract semantic information and then upsamples them to provide dense predictions. UNet \cite{ronneberger2015u} extends it to a U-shape architecture with lateral connections between the downsampling and upsampling path. This architecture and its variants become dominant in medical image segmentation \cite{chen_transunet_2021, unet++, unet3+, nnunet}. The encoder-decoder structure enlarges the receptive field making the Convolution Neural Network (CNN) better at capturing semantic information. Its pyramid structure enables the model to have multi-scale perception and reduces computation complexity. However, the reduction of resolution inevitably loses information, so maintaining semantic information while recovering the spatial resolution becomes challenging. To resolve this issue, multiple upsample techniques \cite{Deconvolutionalnetworks, AdaptiveDeconv, zeiler2014visualizing} have been proposed. However, existing upsample techniques leverage little information from downsampling path.

The prosper of Transformer in the field of Natural Language Processing (NLP) inspires researchers to explore its applicability to Computer Vision (CV). ViT \cite{dosovitskiy2021an} takes only the encoder of the transformer and obtains comparable results as CNN. Swin Transformer \cite{liu2021swin} adopts and modifies the ViT architecture \cite{dosovitskiy2021an} into the one that constructs a hierarchical representation with reduced computation. These works prove the adaptability of pure Transformer to CV downstream tasks such as object detection and segmentation which requires modeling over multi-scale objects and dense pixels. Interestingly, we notice that transformer also possess an encoder and decoder. So, while most researchers focus on the encoder and explore its feature extracting ability, we instead look at the idea of the decoder in transformer and its applicability to segmentation architectures.

A typical decoder in transformer takes the input token embedding of the last position to generate query and obtains the output from encoder to produce key and value \cite{10.5555/3295222.3295349}. Given the circumstances of translation, the output of the decoder is conditioned on the last output tokens while also paying attention to the input sequence tokens. Intuitively, we can view this decoding process where the output of encoders is decoded conditioned on input token embedding. Notice that, the input token sequence may not be as long as the embedding from the encoder. Consequently, if the former is longer, the decoder outputs longer embedding. In a way, we can view it as being upsampled. Inspired by the above analogy, we propose a novel upsample approach, Window Attention Upsample (WAU), which upsamples features conditioned on local and detailed information from the downsampling path in local windows. Attention upsample can enrich the semantic information based on spatial and local information and still outputs features of desired larger shapes.
Considering that large feature maps are unaffordable in global attention, we propose Window Attention Decoder (WAD) to trade-off between the global attention and computation expense. To further ease the learning, we use bilinear upsample to form a residual connection. To the best of our knowledge, we are the first to utilize the transformer decoder in segmentation upsample and explore its ability to upsample feature maps and restore information. We evaluate our method on classic U-Net structure with lateral connection and achieve state-of-the-arts performance on Synapse multi-organ segmentation, Medical Segmentation Decathlon (MSD) Brain and Automatic Cardiac Diagnosis Challenge (ACDC) datasets. We also validate our method on multiple classic architectures and achieve consistent improvement.

In a nutshell, contributions of our work can be summarized as follows:
\begin{itemize}
    \item We propose the idea of upsampling images using the transformer decoder and provide an effective U-shaped architecture for medical image segmentation.
    \item We adopt window-based self-attention to better model pixel-level information and reduce computational cost and memory usage. To further exploit the potential, convolution projection is raised to model locality and residual connection through bilinear interpolation to complement the upsampled feature maps.
    \item Extensive experiments on different datasets using various architectures prove the effectiveness and the generalization ability of our Window Attention Upsample method.
\end{itemize}

\section{Related Work}
FCN \cite{long2015fully} introduces the encoder-decoder architecture and successfully boosts performance in the field of segmentation by a large margin. U-Net \cite{ronneberger2015u} builds upon the idea of FCN and introduces a U-shape network with lateral connections between the downsampling and upsampling path which propagate context information to better localize. Since then, U-shape architecture thrives in many later works of 2D image segmentation \cite{10.1007/978-3-030-00889-5_1,unet3+} and 3D image segmentation \cite{cciccek20163d, milletari2016v}.

Upsample is widely used in semantic segmentation to restore the low resolution feature maps obtained from downsampling path. Conventionally, Interpolation (nearest, bilinear, and cubic) is adopted for the reconstruction of pixels in image processing with each point generated based on its neighbor pixels. Nearest interpolation generates directly from the nearest pixel. Bilinear interpolation \cite{ARCE2005109} estimates a pixel value by averaging the two nearest pixels while cubic \cite{keys1981cubic} evaluate the values of neighbor volumes. Transposed convolution \cite{long2015fully} is proposed to learn an upsample strategy in an end-to-end manner through backpropagation. Besides, latter works including PixelShuffle \cite{shi2016real}, Dupsampling \cite{tian2019decoders}, Meta-Upscale \cite{Hu_2019_CVPR} and CAPAFE \cite{wang2019carafe} are also later development of upsampling techniques.

Attention mechanisms have long been proved useful both in the field of CV and NLP.
SENet \cite{hu2018squeeze} boosts the performance by weighting each channel before it outputs to the next layer. Non-local \cite{wang2018non,wang2020non,he2019non} utilizes pixel-level global attention and models the long-range and global dependencies between pixels. However, global attention at low level layers with large feature maps is impractical for a quadratic complexity with respect to token number \cite{liu2021swin}, thus, Non-local only performs pixel-level attention on low resolution feature maps (\eg the last layer). Our work also models attention upon pixels. To reduce computation, we trade-off between global and local attention by using window attention \cite{liu2021swin}.

Transformer was first introduced in \cite{10.5555/3295222.3295349} for machine translation and since becomes the dominant method in many NLP tasks \cite{devlin-etal-2019-bert,lewis-etal-2020-bart}. Recent works starting with ViT \cite{dosovitskiy2020image} prove the transformer's adaptability in CV. ViT models 16$\times$16 patches of an image as token input to a pure transformer. Swin Transformer leverages local window and shift operation to trade-off between computation and performance. Later works propose numerous techniques that could maintain a reasonable computation budget without overly sacrificing performance \cite{dong2022cswin, ho2019axial, huang2019ccnet, roy-etal-2021-efficient}. Notice that despite our implementation uses local window, these state-of-the-art techniques for sparse and efficient attention can also be incorporated into our method. CvT introduces convolution in the modeling of tokens \cite{wu2021cvt} by leveraging a convolution layer before the standard self-attention layer. Different from their approach, we leverage convolution projection in the self-attention layer to project query, key and value respectively. This incorporates the locality prior into the model and in the meantime reduces computation and parameters (fully connected layer is much larger than a convolution layer and uses more memory when there are numerous tokens).

There are also some recent work demonstrating the transformer's adaptability in medical
image segmentation \cite{dai_transmed_2021,wang_transbts_2021,petit_u-net_2021,chen_transunet_2021}. TransUNet uses a U-shape encoder-decoder architecture. This work exploits the feature extracting ability of ViT and adopts the upsampling path from regular UNet structure where lateral connection passes on local and detailed features for better localization. Furthermore, Hatamizadeh \etal \cite{hatamizadeh_unetr_2021} propose UNETR using solely transformer to extract 3D features. In this work, the transformer encoder proves to be good at modeling long dependency over a 3D input sequence of images. Karimi \etal \cite{karimi_convolution-free_2021} introduce a convolution-free model which utilize solely the transformer as a feature extractor. Given a 3D image block, the corresponding embedding of each patches is computed and the segmentation map is generated according to the correlation between patches via self-attention mechanism. This work is based entirely on transformer without using convolution, which further promotes the application of transformer in medical image processing.

\section{Method}

\subsection{Decoder to Upsample}
Decoder adopts the idea of dot product attention, much like encoder prevalent in recent work of transformer in vision. Unlike the patch encoder, our decoder attention acts on pixel-level instead of patch level in order to better model the dense information. So here, we refer to one pixel as one token.

\begin{figure*}[htbp]
\centering
\includegraphics[width=\linewidth]{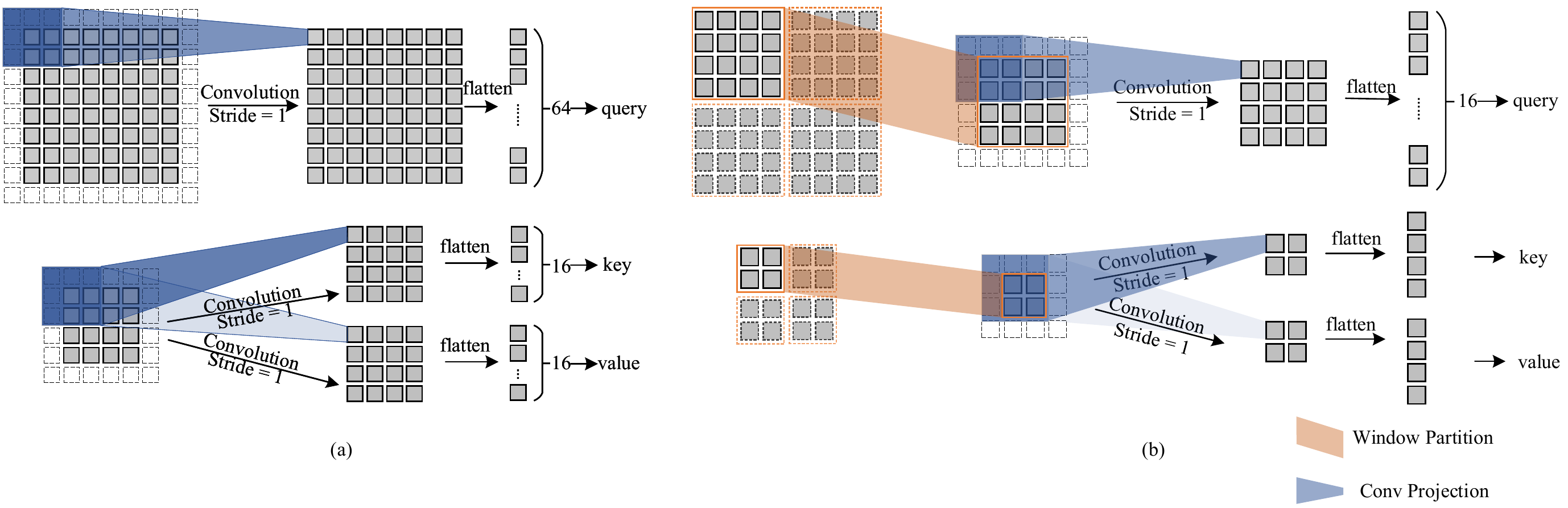}
\centering
\caption{Demonstration of convolution projection in (a) Attention Decoder and (b) Window Attention Decoder with $W=H=4$, $n = 2$ and $M = 4$.}
\label{fig:AD_and_WAD}
\end{figure*}

For the purpose of upsampling, we are majorly concerned about two factors: whether it can maintain or even enrich semantic information necessary for segmentation and whether it outputs feature maps of higher resolution. Transformer decoder inherently uses additional information (\ie query token) to instruct the process of attention by imposing a larger weighting on tokens whose key are similar with query and a smaller weighting otherwise. In our Attention Decoder (AD), we use the feature maps of larger resolution from downsampling path to generate query and input features from upsampling path to generate key and value. In this way, larger resolution feature can be generated conditioned on rich information from downsampling path. This can be formulated as below:
\begin{equation} \hat{z}^l = AD(LN(\hat{z}^{(l-1)}),  LN(\hat{a}^{(l)})) + \hat{z}^{l - 1} \end{equation} where $LN(\cdot)$ represents layer normalization, $\hat{z}^{l - 1} \in  \mathbb{R}^{H^{l - 1} \times W^{l - 1} \times C^{l - 1}}$ denotes features of layer ${l-1}$ in upsampling path and $\hat{a}^{(l)}\in  \mathbb{R}^{H^l \times W^l \times C^l} $ denotes the corresponding feature maps from downsampling path.
\begin{equation} H^{l} = n \cdot H^{l - 1}, W^{l} = n \cdot W^{l - 1} \end{equation} where $H^l, W^l$ denotes the height and width of the feature map and $n$ an integer larger than 1. By taking the context information from the downsampling path, decoder manages to model the global semantic information conditioned on corresponding low level features. Intuitively, context information will increase the weighting of relevant tokens that benefits the upsampling, so the semantic information from upsampling path can be maintained and even enriched.

\subsection{Locality and Computation Considerations}
Locality is an excellent property of CNN, which helps model the local features such as edges and corners. The reconstruction of higher resolution should focus more on neighboring regions. However, the transformer attends to all tokens deprived of the this good property. Despite its ability to model long-range dependency, transformer may lose focus on the significant and relevant tokens when there are numerous tokens, which is an essential problem in the pixel upsample process. Moreover, global attention among all tokens possess a quadratic complexity and memory usage with respect to the number of tokens, which is unaffordable for modeling pixel-level attention, especially at upper layers where resolution is high. To restrict the model's attention in the local area and to reduce computation overhead, we propose to leverage convolution projection and local window attention as detailed in the following sections. 

\subsubsection{Introducing convolution to projection}
\label{sec:Introducing_convolution_to_projection}
To better model local information, we try to incorporate convolution into projection prior to attention block. As shown in Figure \ref{fig:AD_and_WAD}, we use a kernel of size larger than 1, typically 3 to replace the linear projection that is widely used in transformer attention block. In our paper, all convolutions use kernel of 3$\times$3 and maintains sizes (\ie "same" padding).
After the projection, three matrices, key ($k$), value ($v$) and query ($q$) are obtained and then flattened into 1D for subsequent multi-head attention process. Notice, since our input feature maps for query are larger than that of key and value, 1D query sequence are longer than key and value sequence. The output of decoder is the same size as query. After reshaping the output back to 2D, the resolution of the output are the same with feature maps from downsampling path. In this way, upsampling is done. The convolution projection can be written as follows:
\begin{equation} \hat{z_i}^{q} = F(s_c^{q} * LN(\hat{a_i})) \end{equation}
\begin{equation} \hat{z_i}^{k/v} = F(s_c^{k/v} * LN(\hat{z_i}))\end{equation} 
Here * denotes the convolution operator,  $s_c$ = $[s_c^1, s_c^2, \cdots, s_c^{C'}]$ where $C'$ is the number of output channels. $\hat{z_i}^{q/k/v}$ is the corresponding $k$, $q$, $v$ matrices obtained and $F(\cdot)$ denotes an operation that flattens 2D images into 1D sequence. Then we apply dot attention on $k$, $q$, $v$ and computes the upsampled feature maps:
\begin{equation} \label{eq:upsample} \hat{z}^l = s_c' *reshape(softmax(\frac{\hat{z_i}^{q}{\hat{z_i}^{kT}}}{\sqrt{d_k}}) \hat{z_i}^{v})\end{equation} Here, $reshape(\cdot)$ denotes an operation that reshapes the 1D sequence back to 2D feature maps. Another convolution with kernel $s_c'$ is applied after the attention function.

\subsubsection{Attention in Local Window}
Inspired by \cite{liu2021swin}, we propose local window attention for the attention decoder. Since self-attention works on one group of tokens, one window is enough. However, in WAD, we have tokens from two different resolution feature maps, so windows with different sizes are required to align the output key, value and query. As shown in Figure \ref{fig:AD_and_WAD}, we apply windows with different sizes to feature maps from lateral connection and tokens from upsampling path. Inherit from the preceding formulation, feature map from lateral connection $\hat{a}^{(l)}\in  \mathbb{R}^{H^l \times W^l \times C^l} $ is $n$ times the size of that from upsampling path $\hat{z}^{l - 1} \in  \mathbb{R}^{H^{l - 1} \times W^{l - 1} \times C^{l - 1}}$.
In order to align the number windows in query and key, value, windows sizes ratio between the two should also be $n$. With window attention, our WAD can be formulated as below:

\begin{equation} \hat{z}^l = WAD(LN(\hat{z}^{(l-1)}),  LN(\hat{a}^{(l)}))\end{equation} In computational aspect,
suppose we have feature maps $\hat{a}\in  \mathbb{R}^{H_1 \times W_1 \times C} $ from lateral connection  and $\hat{z}^ \in  \mathbb{R}^{H_{2} \times W_{2} \times C}$ from upsampling path , where $H_1 = n \cdot H_2 $, $W_1 = n \cdot W_2 $. For WAD, we use window size of $M_1, M_2$ for $\hat{a},\hat{z}$ respectively, where $M_1 = n \cdot M_2 $.
\begin{equation}   \Omega(AD) = 2H_2W_2C^2k^2(n^2 + 1) + 2(H_2W_2)^2Cn^2 \end{equation}
\begin{equation}\Omega(WAD) = 2H_2W_2C^2k^2(n^2 + 1) +  2(H_2W_2)Cn^2M_2^2\end{equation}where $k$ is the kernel size for our convolution projection. We show here that, with large $H_2, W_2$, AD is generally impractical for a quadratic computation complexity with respect to $H_2W_2$ while WAD is linear to $H_2W_2$ with some fixed $M_2$ and $n$.
As for memory consideration, we have the following:
\begin{equation}
\Omega(AD) = H_2W_2C(n^2 + 2) + n^2(H_2W_2)^2 \end{equation}
\begin{equation}
\Omega(WAD) = H_2W_2C(n^2 + 2) + n^2{M_2}^2H_2W_2
\end{equation}
Notice that the above is the memory usage of intermediate matrices (\ie $k$, $q$, $v$ matrices and attention weights). We show that AD without window attention occupies quadratic memory with respect to $H_2W_2$ while WAD is linear.
With a hyper-parameter M, the method shows great scalability. Given any specific tasks, one can adjust the window size M for a better performance provided limited computation and memory resources.

\subsubsection{Discussion}
Swin transformer leverages the local window attention to save computation resources. Despite its low computation overhead, window attention limits the model's long-range dependency and leads to a degraded performance \cite{liu2021swin}. That's to say, larger window sizes generally leads to better performance \cite{li2022exploring}. To compensate for the loss of long-range dependency, Swin leverages shifted operation to increase the attention range. However, in this work, we discover that when attending to a large number of tokens (at pixel-level), the attention mechanism loses its focus and pays attention to irrelevant parts of the feature map \cite{yang2021focal} as we observe a drop in performance when using larger window sizes in Figure \ref{fig:window_size}. To restrict attention in local areas, which is important for upsampling, window attention is used to confine the attention in local windows. We also conduct an ablation study by adding an additional shifted window attention layer before our upsampling module and observe an even lower performance (72.34 DSC compared with 73.65 DSC). This further demonstrates that simply enlarging receptive fields may be sub-optimal for upsampling.

\subsection{Residual Connection Through Bilinear}\label{sec:residual}
In order to complement the features and form a residual-like operation, we propose to use bilinear interpolation to upsample and adds the two upsampled features together as output. This bilinear upsampled feature serves as a supplement as well as a residual connection that ease the training of WAD.
\begin{equation}
 \hat{z}^l = WAD(LN(\hat{z}^{(l-1)}),  LN(\hat{a}^{(l)})) + Bilinear(\hat{z}^{(l-1)})
\end{equation} 
where $\hat{z}^l$ is the output feature map of decoder upsample module $l$ and $\hat{a}^{(l)}$ are corresponding feature maps of twice the resolution from downsampling path.

\begin{figure}[htbp]
\centering
\includegraphics[width=\linewidth]{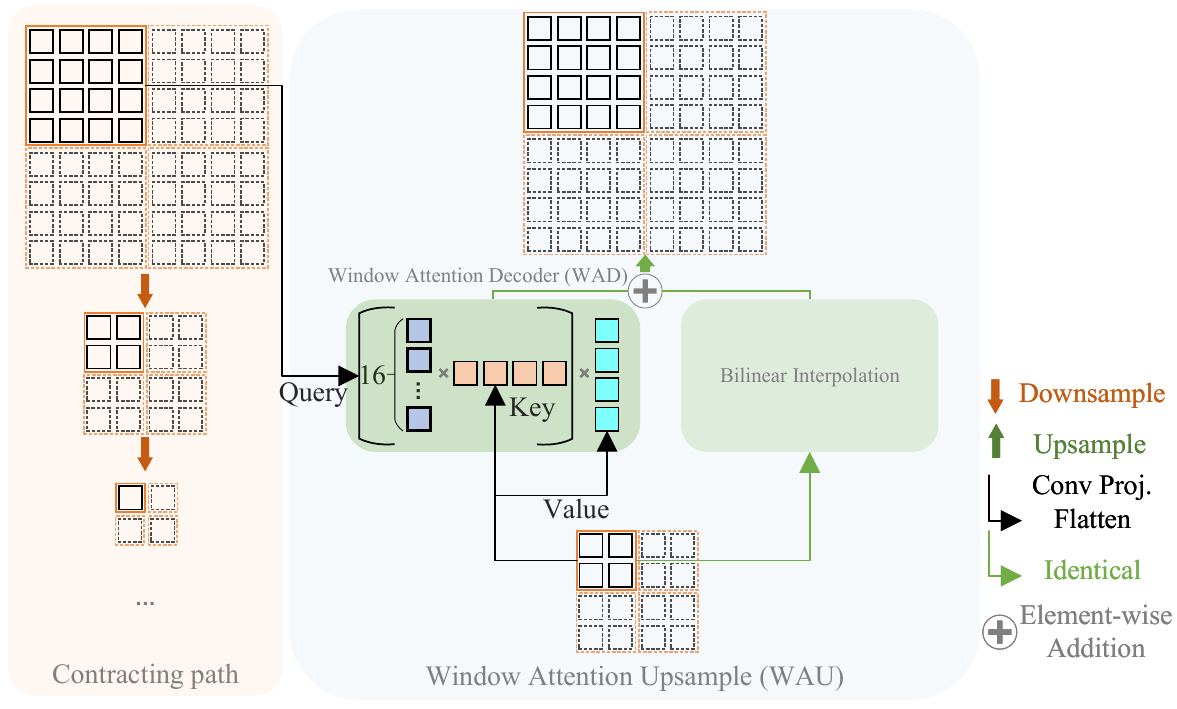}
\caption{Demonstration of WAU with $W=H=4$, $n=2$ and $M=4$. WAD leverages features of larger resolution from downsampling path and features from upsampling path to generate query and key/value respectively, as the embedding of query is longer than the embedding of key/value, the features are then upsampled. Outputs of WAD and Bilinear interpolation are element-wise added to generate upsampled features.}
\label{fig:WAU}
\end{figure}

\subsection{Window Attention Upsample}
Combining ideas from the above, we have Window Attention Upsample (WAU). As shown in Figure \ref{fig:WAU}, WAU possesses two branches, the Window Attention Decoder branch and Bilinear Interpolation branch. Each window of pixels are passed from lateral connection as query and corresponding window from upsampling path serves as key and value. Dot attention is performed on key and query to compute attention weights. The final output of such window is obtained by multiplying the attention weights and the value matrix. All windows are computed simultaneously to form a larger feature map. After both WAD and Bilinear Interpolation is done, the results of the two are summed as the final output.

\subsection{Instantiation}
To evaluate the effectiveness of our upsample method, we incorporate our method into several classic and state-of-the-art network architectures such as ResUNet, 3D-UNet, FCN, and DeepLabV3. Generally, to incorporate our method into an existing architecture, we simply replace the original upsample layer with WAU. Since WAU requires the feature map from the downsampling path, we build up lateral connections to pass the dowsampled feature maps of the desired shape to each of the WAU module. Take ResUNet as an example, we replace all of the original bilinear upsample modules with WAU. Then, we leverage the lateral connection which feeds a feature map of exactly twice the size. Consequently, each WAD upsamples the feature map twice the size and progressively upsamples the feature map to the original input sizes. We present an example of instantiations where the classic UNet is combined with WAU, as shown in Figure \ref{fig:architeture}. Detailed architecture description of each instantiation is provided in Appendix \ref{SecondAppendix}.

\begin{figure*}[htbp]
    \centering
    \includegraphics[width=0.8\linewidth]{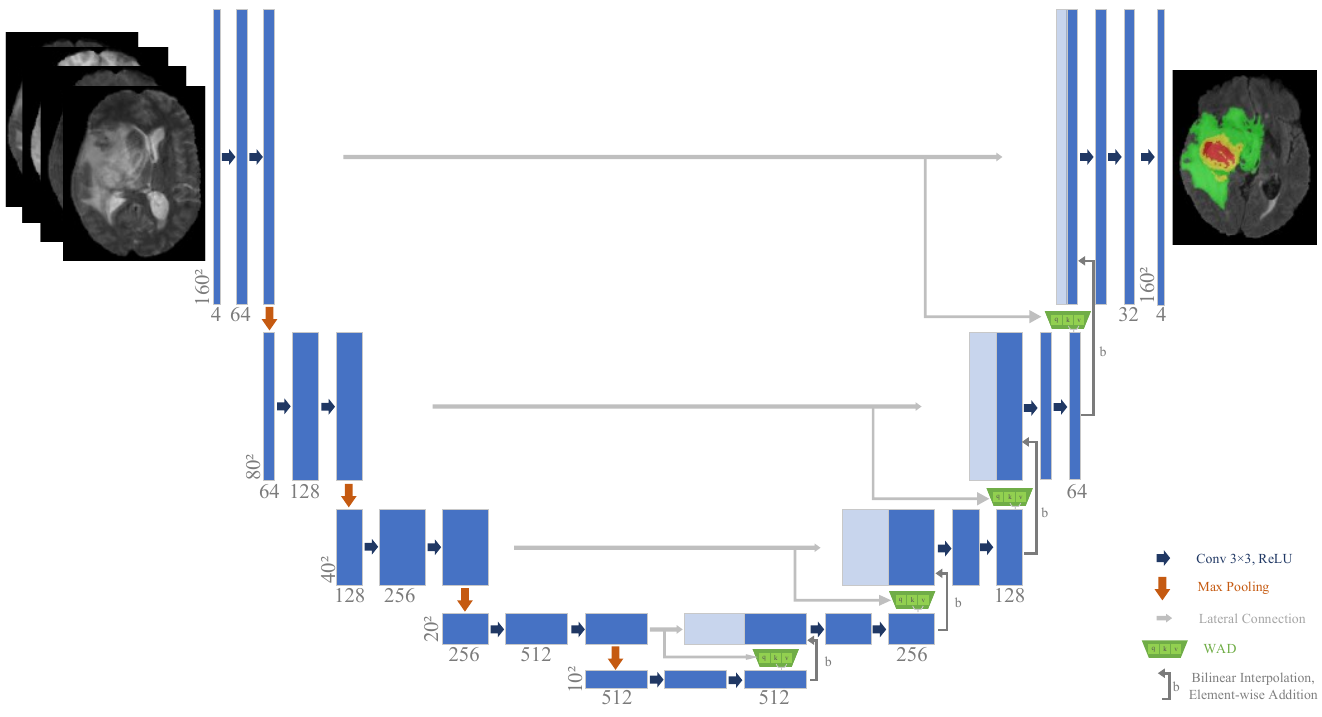}
    \caption{An example of instantiations where the classic UNet is combined with WAU. Input image is first downsampled 4 times, then upsampled through our proposed WAU module. At each WAU, queries from encoding path are passed through lateral connection to query the feature maps (processed as keys) from below.}
    \label{fig:architeture}
\end{figure*}

\section{Experiment}

\subsection{Dataset}
We evaluate our model on the MSD Task01 BrainTumour dataset (MSD Brain) \cite{simpson2019large}, Synapse multi-organ segmentation dataset (Synapse), and Automatic Cardiac Diagnosis Challenge (ACDC) datasets. MSD Brain contains 484 multimodal multisite MRI data (FLAIR, T1w, T1gd, T2w) and four labels including background, edema (Ed), non-enhancing tumor (NET), and enhancing tumor (ET). For MSD Brain, we apply z-scoring normalization to preprocess each case. To alleviate the problem of class imbalance, we remove all blank slices with zero values and crop each slice to the region of nonzero values. Each slice is cropped to 128$\times$128 before feeding into the model. Synapse contains 30 cases with a total of 3779 slices of resolution 512$\times$512. Each case consists of 14 labeled organs from MICCAI 2015 Multi-Atlas Abdomen Labeling Challenge. Following the settings of TransUNet \cite{chen_transunet_2021}, we select 8 organs for model evaluation and divide cases into train set and validation set with the ratio of 3:2 (\ie 18 cases for training and 12 for validation). Preprocess pipeline includes clipping the values of each pixel to [-125, 275] and normalizing them to [0, 1]. Both datasets are trained on 2D slices and validated on 3D volume following standard evaluation procedure. ACDC contains 100 cases of MRI scans from different patients whose goal is to segment the myocardium (Myo) of the left ventricle and the cavity of the right (RV) and left ventricle (LV). Following the settings of \cite{10.1007/978-3-319-75541-0_12}, 80 and 20 subjects are divided into training and validation set respectively with the resolution re-sampled to 160$\times$160. Multiple data preprocessing techniques are done with the same settings of \cite{pmlr-v121-nguyen20a}.

\subsection{Implementation Details}
\label{sec:Implementation_Details}
For all experiments, we perform some slight data augmentation, \eg, random rotation, and horizontal and vertical flipping. For model invariant, to coincide with the typical U-Net structure, we set $n = 2$ meaning to upsample by 2 at each WAU module. We use a base window size of 4 for the MSD Brain dataset and 7 for the Synapse dataset. All models are trained using Adam \cite{DBLP:journals/corr/KingmaB14} with betas of 0.9 and 0.999 (default setting) and Cosine Annealing learning rate \cite{DBLP:conf/iclr/LoshchilovH17} with a warm up \cite{he2016deep} of 2 epochs. The initial learning rate is 0.0001 with a batch size of 12 for MSD Brain and 32 for Synapse. No pre-training is used and all experiments are conducted using two NVIDIA RTX2080Ti GPU.

\subsection{Results}
\subsubsection{Results on MSD Brain Dataset}
\label{sec:Results_on_MSD_Brain_Dataset}
Results of our model and other state-of-the-art methods are shown in Table \ref{tab:msdbrain} \footnote{Results of UNet, VNet, AHNet, Att-UNet, SegResNet and UNETR are from \cite{hatamizadeh_unetr_2021}, results of two nnUNet models are from \cite{nnunet}, results of DiNTS are from \cite{he2021dints}.}. On the MSD BrainTumour Dataset, our model achieves the best performance of 74.75\% Dice similarity coefficient (DSC) with 80.73\%, 63.23\%, and 80.29\% on edema, non-enhancing tumor, and enhancing tumor respectively. When comparing with our baseline model ResUNet, we achieve a significant increase of 2.83\%. Compared with nnUNet \cite{nnunet} 2D version, which also builds upon the U-Net architecture, our method obtains an improvement of 3.19\%. Moreover, when compared with ensemble 3D nnUNet, we also outperform by 0.86\%. We also make a comparison between state-of-art Transformer-based models including recent 3D network UNETR \cite{hatamizadeh_unetr_2021} and 2D network SwinUNet \cite{cao2021swin} which we outperform by a margin of 2.94\% and 1.55\% on average DSC respectively. To provide a demonstration of results on MSD Brain dataset, the first two rows of Figure \ref{fig:results} offer a sample of (a) gt label, (b) ResUNet, (c) TransUNet, and (d) Ours. Our baseline ResUNet model shows to be under-segmented prone, \ie the first row of (b) shows an incomplete segmentation region while transformer-based models, \ie TransUNet and Ours, can produce more complete and accurate results via the establishment of long-range dependencies. We can also see that both ResUNet and TransUNet face the problem of providing false positive predictions, \ie the second row (b) and (c) show a false positive prediction of ET instead of NET (gt). Compared with TransUnet, our model shows great performance in local and marginal regions. This could be attributed to pixel-level correlation in the local window that could better model the local features.

\begin{table}[htbp]
\centering
\caption{Comparison with State-of-the-art on the MSD Brain dataset.
}
\scalebox{1}{
\setlength{\tabcolsep}{3mm}{
\begin{tabular}{l|c|ccc}
\hline
Methods                               & DSC $\uparrow$ & Ed             & NET           & ET            \\ \hline
FCN32s \cite{long2015fully}          & 60.40           & 70.03           & 46.96          & 64.99          \\
FCN16s \cite{long2015fully}          & 66.25          & 74.84           & 52.93          & 70.97          \\
FCN8s \cite{long2015fully}           & 69.21          & 76.61           & 56.17          & 74.83          \\
UNet \cite{ronneberger2015u}         & 67.65          & 75.03           & 57.87          & 70.06          \\
DeepLabV3 \cite{chen2017rethinking}  & 68.86          & 77.42           & 57.11          & 72.04          \\
TransUNet \cite{chen_transunet_2021} & 71.11          & 77.38           & 59.04          & 76.91          \\
nnUNet (2D) \cite{nnunet}           & 71.56          & 78.60           & 58.65          & 77.42          \\
TransBTS \cite{wang2021transbts} & 71.79 & 78.62 & 60.14 & 76.61
\\
ResUNet                              & 71.92          & 77.73           & 59.47          & 78.57          \\
SegTran R50 \cite{li2021medical} & 73.48 & 80.20 & 61.81 & 78.42
\\
SwinUNet \cite{cao2021swin}         & 73.20          & 79.41           & 61.38          & 73.20          \\ \hline
VNet \cite{milletari2016v}           & 65.77          & 75.96           & 54.99          & 66.38          \\
AHNet \cite{liu20183d}               & 66.63          & 75.8            & 57.58          & 66.50          \\
Att-UNet \cite{oktay2018attention}   & 67.07          & 75.29           & 57.11          & 68.81          \\
SegResNet \cite{myronenko20183d}     & 69.65          & 76.37           & 59.56          & 73.03          \\
UNETR \cite{hatamizadeh_unetr_2021}  & 71.81          & 79.00           & 60.62          & 75.82          \\
DiNTS \cite{he2021dints} & 72.97 & 80.20 & 61.09 & 77.63
\\
3D-UNet \cite{cciccek20163d}         & 72.15          & 79.45           & 60.42          & 76.59          \\
nnUNet (3D) \cite{nnunet}           & 73.89          & \textbf{80.79} & 61.72          & 79.16          \\ \hline
Ours                                 & \textbf{74.75} & 80.73           & \textbf{63.23} & \textbf{80.29} \\ \hline
\end{tabular}
}}
\label{tab:msdbrain}
\end{table}

\begin{table*}[htbp]
\centering
\caption{Comparison with State-of-the-art on the Synapse dataset.
}
\setlength{\tabcolsep}{0.3mm}{

\resizebox{0.7\linewidth}{!}{
\begin{tabular}{l|cc|cccccccc}
\hline
Methods      & DSC $\uparrow$          & HD $\downarrow$            & Aorta          & GB    & Kidney(L)      & Kidney(R)      & Liver          & Pancreas       & Spleen         & Stomach        \\
\hline
V-Net \cite{milletari2016v}        & 68.81          & -              & 75.34          & 51.87          & 77.10          & 80.75 & 87.84          & 40.05          & 80.56          & 56.98          \\
DARR \cite{fu2020domain}         & 69.77          & -              & 74.74          & 53.77          & 72.31          & 73.24          & 94.08          & 54.18          & 89.90 & 45.96          \\
R50 U-Net \cite{chen_transunet_2021}    & 74.68          & 36.87          & 87.74          & 63.66          & 80.60          & 78.19          & 93.74          & 56.90          & 85.87          & 74.16          \\
R50 Att-UNet \cite{chen_transunet_2021} & 75.57          & 36.97          & 55.92          & 63.91          & 79.20          & 72.71          & 93.56          & 49.37          & 87.19          & 74.95          \\
R50 ViT \cite{chen_transunet_2021}      & 71.29          & 32.87          & 73.73          & 55.13          & 75.80          & 72.20          & 91.51          & 45.99          & 81.99          & 73.95          \\
TransUNet \cite{chen_transunet_2021}    & 77.48          & 31.69          & 87.23          & 63.13          & 81.87          & 77.02          & 94.08 & 55.86          & 85.08          & 75.62          \\
SwinUNet \cite{cao2021swin} & 79.13& 21.55 & 85.47& \textbf{66.53}& 83.28& 79.61& 94.29& 56.58 &  \textbf{90.66}& 76.60\\
U-Net \cite{ronneberger2015u}        & 73.09          & 40.05          & 83.17          & 58.74          & 80.40          & 73.36          & 93.13          & 45.43          & 83.90          & 66.59          \\
ResUNet & 74.99& 27.57 & \textbf{88.55}& 59.93& 83.14& 71.63& 93.16& 52.51 &  84.23& 66.77\\
\hline
Ours & \textbf{80.40} & \textbf{18.50} & 88.40          & 60.64 & \textbf{84.47} & \textbf{81.04}          & \textbf{94.40}          & \textbf{66.01} & 88.92          & \textbf{79.35}\\
\hline
\end{tabular}
}

\label{tab:synapse}
}
\end{table*}

\begin{table}[htbp]
\centering
\caption{Statistics of model parameters and FLOPs.}
\scalebox{1}{
\setlength{\tabcolsep}{3mm}{
\begin{tabular}{l|cc}
\hline
 Methods      & Params  & GFLOPs \\ \hline
 ResUNet     & 17.27M  & 14.64  \\
                            wide-ResUNet & 30.82M  & 26.08     \\
                            TransUNet   & 93.19M  & 11.71  \\
                            SegTran R50 & 128.82M & 53.47      \\
                            SwinUNet    & 27.12M  & 5.49   \\
                            Ours        & 21.80M  & 15.94  \\ 
                           \hline

\end{tabular}
}}

\label{tab:stat}
\end{table}

\subsubsection{Results on Synapse Dataset}

Experiment on Synapse dataset (Table \ref{tab:synapse}) demonstrates the effectiveness and generalization ability to multi-organ tasks of our upsample method. We make comparison with baseline model ResUNet, recent work TransUNet \cite{chen_transunet_2021} and SwinUNet\cite{cao2021swin} where our method outperforms ResUNet by 5.41\%, TransUNet by 2.92\% and SwinUNet by 1.27\% on average DSC \footnote{Results of V-Net, DARR, R50 U-Net, R50 Att-UNet, R50 ViT and TransUNet are from \cite{chen_transunet_2021}.}. Moreover, we make a comparison on Hausdorff (HD) metrics which measures models' sensitivity to edge segmentation. As per the table, we also achieve a state-of-the-art performance of 18.50. Specifically, We achieve the best performance on Kidney(L) with 84.47\%, Kidney(R) with 81.04\%, Liver with 94.40\%, Pancreas with 61.01\%, and Stomach with 79.35\%. This experiment shows our model's ability to generalize to multiple organs' segmentation. To provide a demonstration of results on the Synapse dataset, the middle two rows of Figure \ref{fig:results} offer a sample of (a) gt label, (b) ResUNet and (c) TransUNet, and (d) Ours. From the graph presented, we can also notice the same problem mentioned in Section \ref{sec:Results_on_MSD_Brain_Dataset}, incomplete prediction compared with gt, \ie in the orange region of the third row, (b) shows no positive prediction and (c) shows little positive predictions, and misclassification of the label, \ie in the green and orange rectangle of forth row, both (b) and (c) make false positive predictions.

\subsubsection{Results on ACDC Dataset}
\begin{figure}[htbp]
\centering
\includegraphics[width=\linewidth]{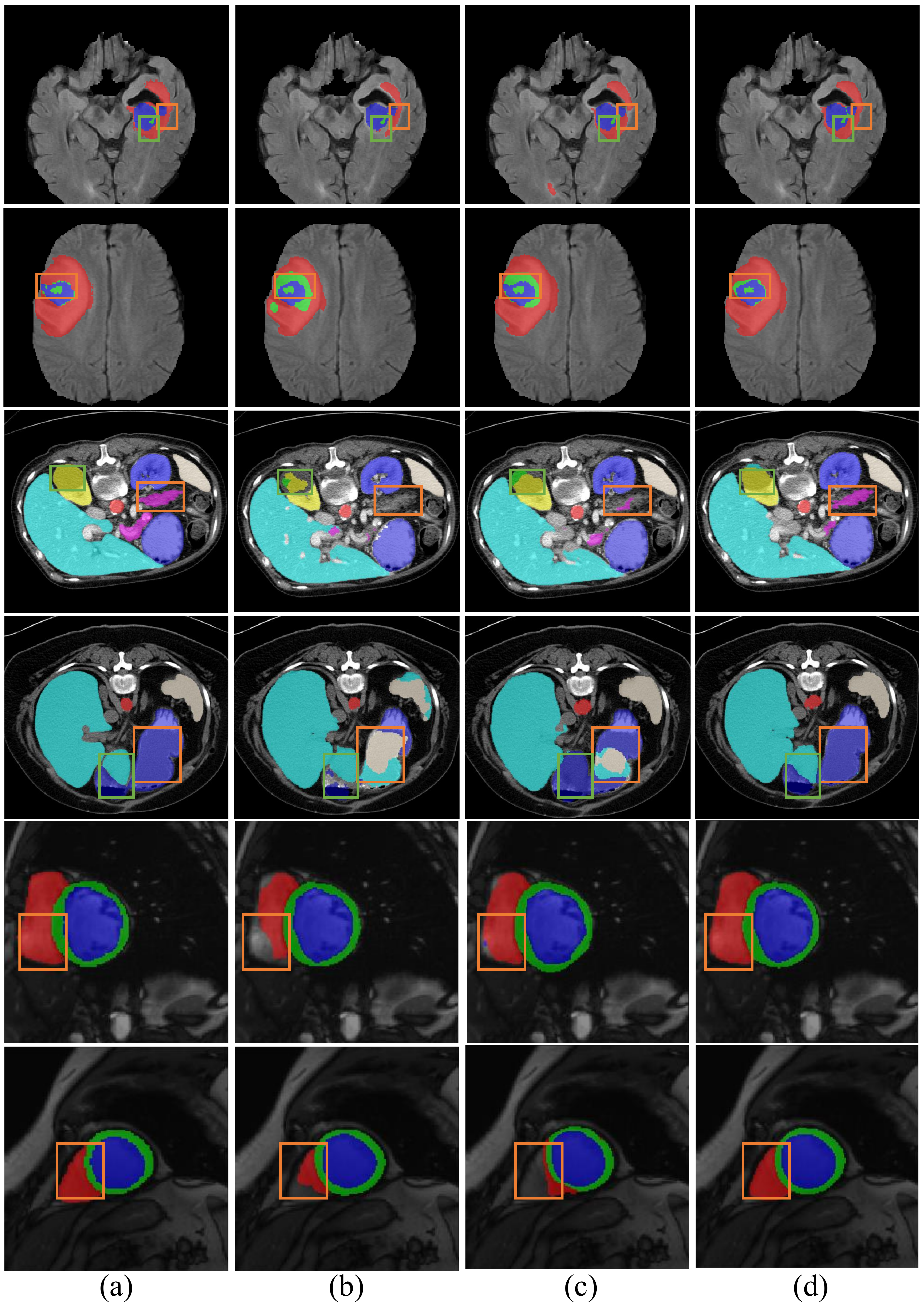}
\centering
\caption{Qualitative results from the MSD Brain, Synapse and ACDC datasets. We compare (a) Ground Truth with the outputs of (b) ResUNet, (c) TransUNet and (d) Ours.}
\label{fig:results}
\end{figure}

Table \ref{tab:acdc} demonstrates our model's performance on ACDC dataset comparing with state-of-the-arts \footnote{Results of R50 U-Net, R50 Att-UNet, ViT-CUP, R50-ViT-CUP, and TransUNet are from \cite{chen_transunet_2021}, results of SwinUNet is from \cite{cao2021swin}.}. On ACDC, we achieve more than 2 points improvement on a benchmark dataset with an average DSC of 90\%+, which we consider quite tremendous on such dataset. It's worth mentioning that after being carefully tuned, ResUNet is capable of achieving a performance of 90.06\%, even higher than other state-of-the-arts such as SwinUNet and TransUNet. However, our model can outperform it by nearly 2 points in DSC. The bottom two rows of Figure \ref{fig:results} offers a sample of gt label(a), ResUNet(b) and TransUNet(c) and Ours(d) from ACDC dataset. We can also observe an incomplete segmentation problem from (b) and (c) of the two rows and our more complete results. This also proves that our method can maintain and even enrich the information in feature maps during upsample process.

\begin{table}[htbp]
\centering
\caption{Comparison with State-of-the-art on the ACDC dataset.
}
\scalebox{1}{
\setlength{\tabcolsep}{3mm}{
\begin{tabular}{l|c|ccc}
\hline
Methods       & DSC $\uparrow$   & RV    & Myo   & LV    \\
\hline
R50-U-Net \cite{chen_transunet_2021}    & 87.55 & 87.10 & 80.63 & 94.92 \\
R50-AttnUNet \cite{chen_transunet_2021} & 86.75 & 87.58 & 79.20 & 93.47 \\
ViT-CUP \cite{chen_transunet_2021}      & 81.45 & 81.46 & 70.71 & 92.18 \\
R50-ViT-CUP \cite{chen_transunet_2021}  & 87.57 & 86.07 & 81.88 & 94.75 \\
TransUNet \cite{chen_transunet_2021}    & 90.05 & 90.14 & 86.00 & 94.00 \\
SwinUNet \cite{cao2021swin}    & 90.00 & 88.55 & 85.62 & \textbf{95.83} \\
ResUNet    & 90.06 & 88.86 & 86.75 & 94.57 \\
\hline
Ours         & \textbf{92.00}    & \textbf{91.65} & \textbf{88.95} & 95.40\\
\hline
\end{tabular}
}}

\label{tab:acdc}
\end{table}

\subsection{Analytical study}

\subsubsection{Comparison with Baseline}
We compare our WAD with different upsample methods including bilinear interpolation, transposed convolution, pixel shuffle \cite{shi2016real} and CARAFE \cite{wang2019carafe}. Table \ref{tab:baseline} shows the performance of various upsample methods on different backbones, we can make the following observations: (i) Despite the difference in upsample method, the overall performance of ResUNet is better than that of classic U-Net, which is why we use ResUNet as the backbone. (ii) Bilinear interpolation is slightly better than the other three upsample methods, but the performance of our proposed WAU far exceeds them all, which suggests that the classic decoder design can be better replaced by our Window Attention Upsample (WAU) strategy. 

It is worth mentioning that we also compare the number of parameters and FLOPs used in ResUNet, TransUNet, SwinUNet, SegTran R50, and our model (Table \ref{tab:stat}). We suppose the better performance of TransUNet over ResUNet could be attributed to the large parameters. Our model, however, uses much fewer parameters (only 1/3 of TransUNet) and relatively acceptable operations to achieve much better performance on all three datasets. To make a fair comparison with baseline ResUNet in terms of parameters, we increase the base channels from 64 to 72, resulting in the wide-ResUNet with more parameters and flops (30.82M and 26.08 GFLOPs). As per the third row of table \ref{tab:baseline}, we can observe that our method still outperforms this improved baseline by more than 2 DSC. This demonstrates that the improvement is not the result of simply adding more parameters and flops. 
\begin{table}[htbp]
\centering
\caption{Comparison of different upsample strategy on MSD Brain dataset.}
\scalebox{1}{
\setlength{\tabcolsep}{1.7mm}{
\begin{tabular}{l|l|c|l}
\hline
Methods                  & Backbone                   & Upsample                        & \multicolumn{1}{c}{DSC $\uparrow$}               \\ \hline
\multirow{11}{*}{UNet}   & \multirow{2}{*}{UNet}      & Bilinear                        & 71.91 (+0.11)                           \\
                         &                            & Transposed                      & 71.80                                            \\ \cline{2-4} 
                         & wide-ResUNet               & Bilinear                        & 72.14                                            \\ \cline{2-4} 
                         & \multirow{6}{*}{ResUNet}   & Bilinear                        & 71.92 (+0.61)                                    \\
                         &                            & Transposed                      & 71.85 (+0.54)                                    \\
                         &                            & pixelShuffle \cite{shi2016real} & 71.31                                            \\
                         &                            & CARAFE \cite{wang2019carafe}    & 71.63 (+0.32)                                    \\ \cline{3-3}
                         &                            & WAD                             & 73.84 (+2.53)                                    \\
                         &                            & WAU (WAD w/ Bilinear)           & \textbf{74.75} (\textbf{+2.83})                  \\ \cline{2-4} 
                         & \multirow{2}{*}{UNet 3D}   & Bilinear                        & 72.15                                            \\
                         &                            & WAU                             & \textbf{72.51} (\textbf{+0.36})                           \\ \hline
\multirow{2}{*}{DeepLab} & \multirow{2}{*}{DeepLabV3} & Bilinear                        & 68.86                                            \\
                         &                            & WAU                             & \textbf{70.33} (\textbf{+1.47}) \\ \hline
\multirow{6}{*}{FCN}     & \multirow{2}{*}{FCN 32s}   & Transposed                      & 60.40                                            \\
                         &                            & WAU                             & \textbf{65.89} (\textbf{+5.49})                           \\ \cline{2-4} 
                         & \multirow{2}{*}{FCN 16s}   & Transposed                      & 66.25                                            \\
                         &                            & WAU                             & \textbf{68.97} (\textbf{+2.72})                           \\ \cline{2-4} 
                         & \multirow{2}{*}{FCN 8s}    & Transposed                      & 69.20                                            \\
                         &                            & WAU                             & \textbf{71.32} (\textbf{+2.12})                           \\ \hline
\end{tabular}
}}

\label{tab:baseline}
\end{table}

\subsubsection{Residual Connection through Bilinear}
We adopt Bilinear Interpolation to form a residual connection. We argue that this process feeds identical mapping forward, and thus can ease the training process. Moreover, the Bilinear Interpolation, in a way, can be viewed as a complement of the upsampled features maps. In this section, we perform an ablation study on this operation. Particularly, we train models with and without bilinear residual connection (\ie, WAU and WAD) on the MSD Brain dataset. From Table \ref{tab:baseline}, we can see that adding Bilinear Interpolation increases DSC by 0.79, which sufficiently proved the effectiveness of residual connection through Bilinear Interpolation.

\begin{figure}[htbp]
\centering
\includegraphics[width=\linewidth]{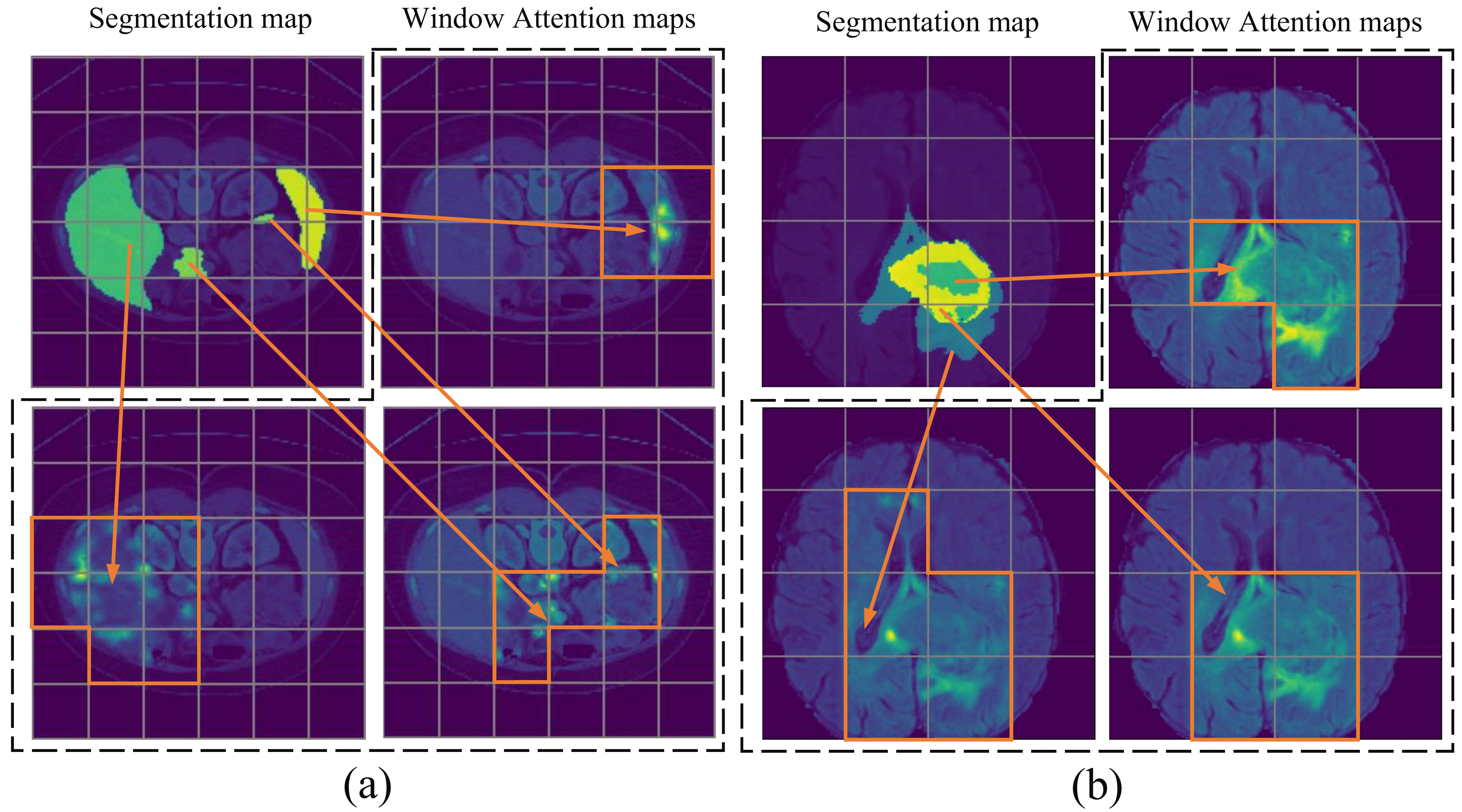}
\centering
    \caption{Visualization of Window Attention weights on (a) Synapse and (b) MSD Brain datasets.}
\label{fig:att_vis}
\end{figure}
\subsubsection{Convolution Matters}
\begin{figure}[htbp]
    \centering
    \includegraphics[width=0.9\linewidth]{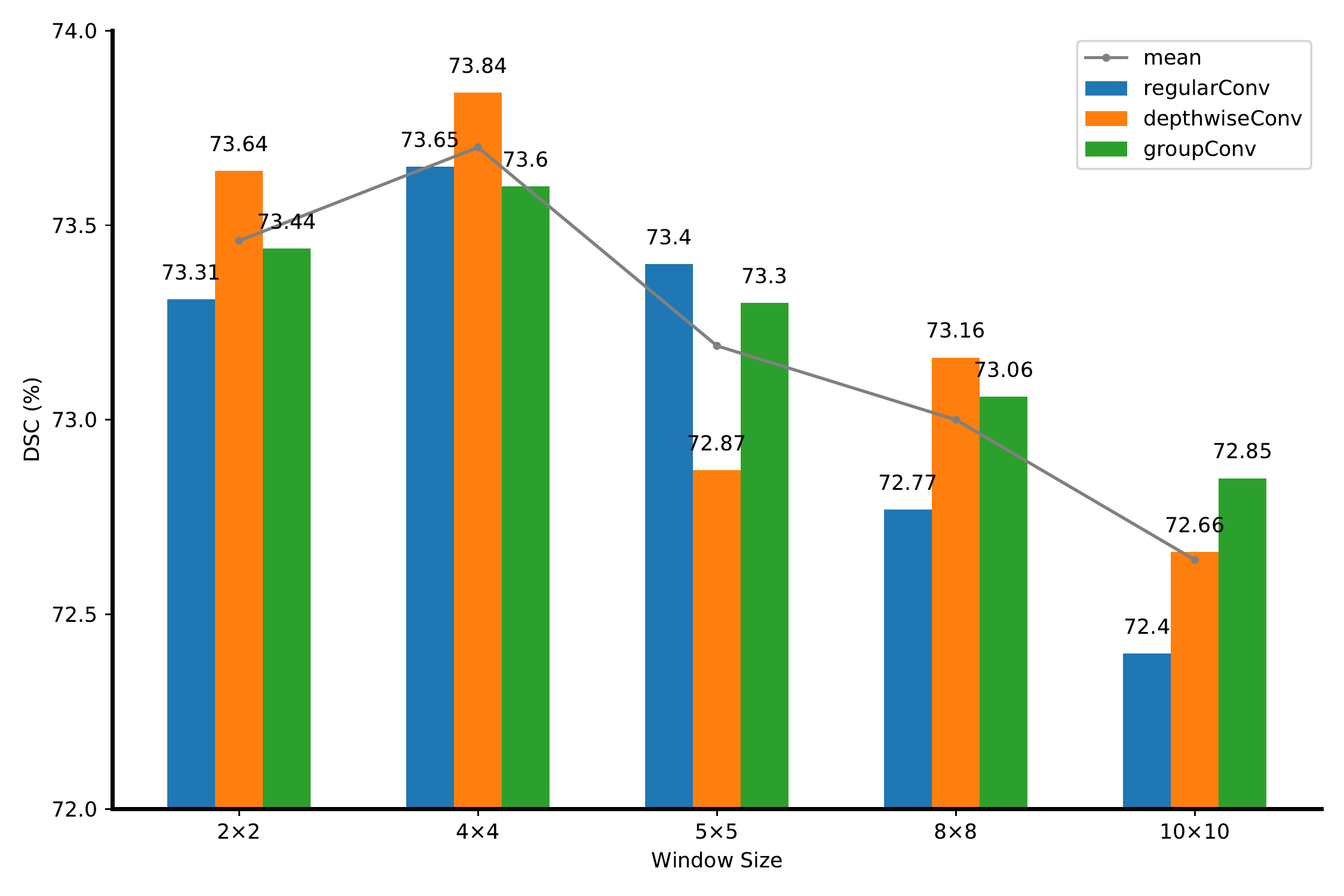}
    \caption{Ablation study on different window sizes and convolution types.}
    \label{fig:window_size}
\end{figure}

In Section \ref{sec:Introducing_convolution_to_projection}, we introduce the convolution projection to obtain the key, query, and value matrices. Compared with linear projection, convolution operation provides modeling on local features which benefits the reconstruction of high-resolution features. In this section, we explore the performance of different convolution operations. In particular, we explore Group Convolution, Depthwise Separable Convolution, and Regular Convolution operation on MSD Brain dataset. Results in Figure \ref{fig:window_size} reveal that Depthwise Separable convolution is slightly better than the other two convolution operations with a window size of 4. This could be attributed to the fact that Depthwise Separable Convolution possesses fewer parameters and thus provides better performance on a relatively small dataset. We also compare the effect of different kernel sizes in Appendix \ref{FirstAppendix}.

\subsubsection{Ablation study on Window Size}
We conduct ablation study on different window sizes with WAD and ResUNet architecture on MSD Brain dataset. As per Figure \ref{fig:window_size}, we find that the optimal window size is 4. Increasing the window size leads to a drop in the performance. We hypothesize that the model might lose focus when the window size is too large and this is particularly problematic for upsampling as it depends on the detailed and local features in neighboring regions. Smaller window is better as it restricts model's attention spatially. However, using too small a window also degrades the performance since it gets rid of the long-range dependency. We can observe a drop in performance when the window size is set to 2.

\subsubsection{Generalizability of WAU with Different Architecture}
We argue that our proposed WAU can be incorporated into any architecture that possesses lateral connections. To prove the generalizability of our proposed WAU, we incorporate our method into different architectures and observe consistent improvements in all experiments. Specifically, we incorporate WAU into UNet 3D and observe an improvement of 0.36 DSC. This demonstrates that our method can be used in 3D volume segmentation which comprises a large category of medical segmentation methods. We also observe an improvement of at most 5 points on the three variants of classic FCN and an improvement of 1.47 on the DeepLabV3 model. Results are displayed in table \ref{tab:baseline}.

\subsection{Visualization}
In this part, we provide visualization of Window Attention weights and the upsampled feature maps of different models in the upsampling path. To obtain our Window Attention weights, we retrieve the attention weights in WAD. Since each attention is computed inside local windows, we select the activated regions (the positive region in ground truth) and show the average attention weights of these windows with positive pixels. Feature maps after every upsample module are also visualized to demonstrate the effectiveness of our method. Figure \ref{fig:att_vis} is the visualization of our Window Attention weights and shows how the attention is focused on the relevant pixels of the target area in each window. This further demonstrates the validation of our methods that by imposing an attention, model is prone to focus more on target. This enriches the information needed for segmentation task and leads to better performance. Figure \ref{fig:decoder_feature} presents the upsampled feature maps after every upsample procedure on MSD Brain and Synapse datasets. It can be seen that our upsample method, taking advantage of self-attention, focuses better on target area than pure CNN-based method (\ie ResUNet). Also, compared with ResUNet, our method shows a clear lesion that could further assist the diagnosis.

\section{Conclusion}
In this paper, we present the first study to explore the adaptability of transformer decoder in segmentation and its usage in upsample. Our work proves that decoder can also be adopted to model visual information and performs even better than traditional upsample techniques. To leverage the ability of such architecture, we propose our Window Attention Upsample that reconstruct semantic pixels to desired shape conditioned on local and detailed information. With this, we provide a better alternative to the basic upsample operation and can be fused in any segmentation model that requires upsample. Moreover, our work partly exploits the possibility of adopting a pure transformer with encoder and decoder into CV.

\section*{Acknowledgment}
This work is jointly supported by Guangdong Provincial Key Laboratory of Artificial Intelligence in Medical Image Analysis and Application (No. 2022B1212010011) and the International Cooperation Project of Guangdong Province under grants 2021A0505030017.

{\small
\bibliographystyle{IEEEtranS}
\bibliography{egbib}
}

\clearpage
\appendices
\section{Ablation study on different kernel sizes}
\label{FirstAppendix}

\begin{figure}[htbp]
    \centering
    \includegraphics[width=0.9\linewidth]{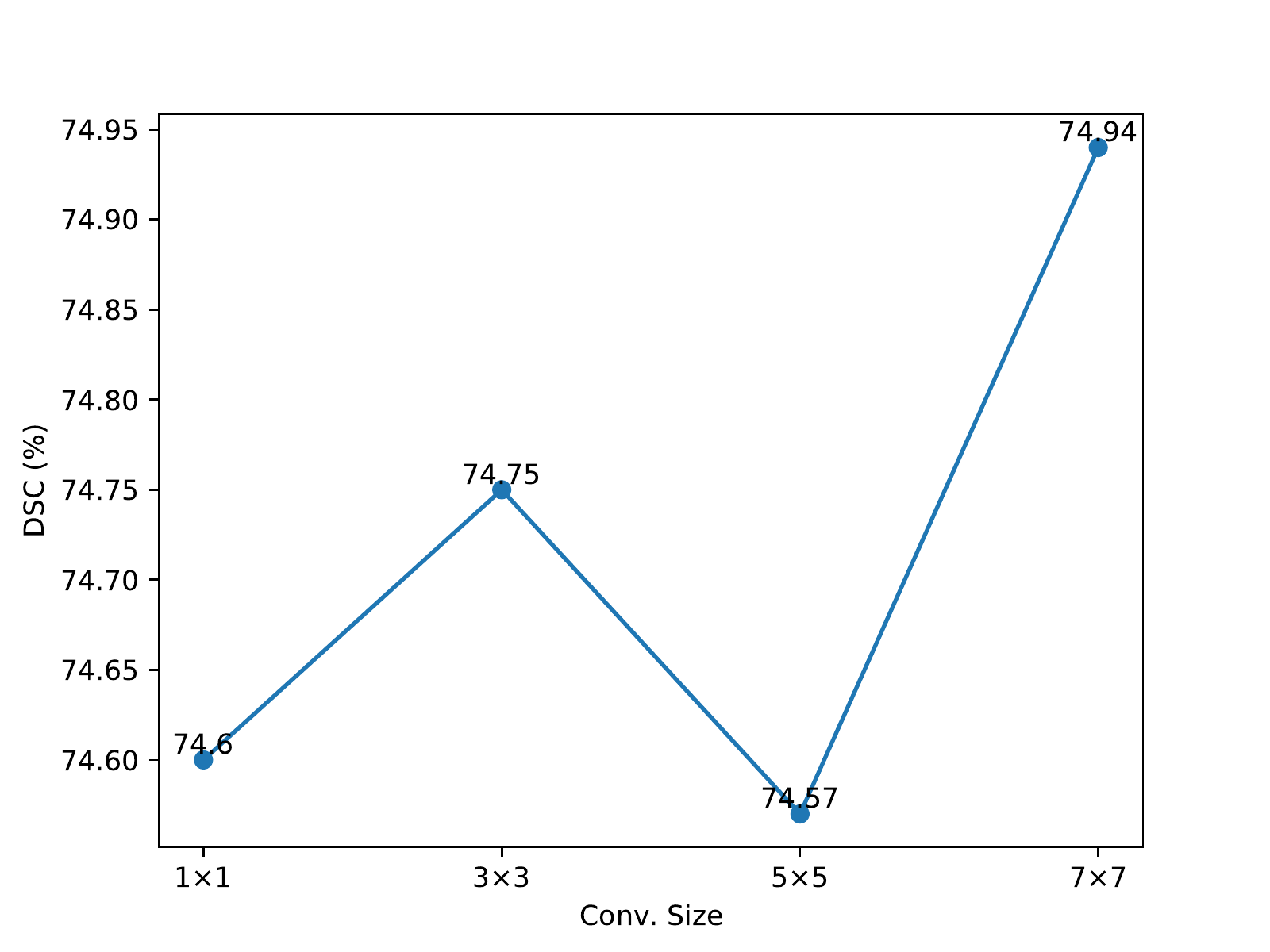}
    \caption{Ablation study on different kernel sizes of convolution projection.}
    \label{fig:conv_size}
\end{figure}
We conduct an ablation study on different kernel sizes using WAU with a base window size of 4 on the MSD Brain dataset. Figure \ref{fig:conv_size} shows the performance of different kernel sizes in convolution projection, where we can find that 7$\times$7 convolution kernels can achieve slightly better performance than the 3$\times$3 convolution kernels. For simplicity, we leverage 3$\times$3 convolution kernels in WAU.

\section{Implementation Details of Different Architectures}
\label{SecondAppendix}

In Table \ref{tab:baseline} we demonstrate the generalizability of our proposed WAU method. Here we provide the details of implementation for incorporating WAU into different architectures.

We first discuss the modifications from UNet to ResUNet. We add a residual connection via an additional 3$\times$3 convolution layer on every two convs block of the original UNet. The rest remains the same as UNet. For the UNet family, including UNet, ResUNet, and UNet 3D, we directly leverage its skip connections from downsampling path to upsampling path to form $qurey$ vector. We use the feature maps in lower resolution from the previous layer to form $key$ and $value$ vectors. According to Equation \ref{eq:upsample}, the low-resolution feature map will be upsampled conditioned on the larger feature map from the downsampling path. The overall architecture is illustrated in Figure \ref{fig:architeture}. 

For the DeepLab family, the $qurey$ vector is acquired from the encoder. We utilize the feature maps with the same resolution as the input images. This is also the expected resolution of the upsampled feature maps. The $key$ and $value$ vectors are formed by the output of the ASPP module. The encoded feature maps will be upsampled only once by 16$\times$ via WAU.

The FCN methods are similar to the DeepLab series. We replace the original 32$\times$, 16$\times$, and 8$\times$ upsampling by WAU with $query$ feature from the encoder. For FCN 32s, we utilize WAU only once and upsample the feature maps to input sizes. For FCN 16s and 8s, multiple WAU modules are inserted to replace the original upsample module. Each WAU leverages the feature maps from a specific layer of the encoder to generate $query$.

\end{document}